\documentclass[10pt,twocolumn,letterpaper]{article}

\usepackage{cvpr}              

\usepackage[dvipsnames]{xcolor}

\definecolor{cvprblue}{rgb}{0.21,0.49,0.74}
\usepackage[pagebackref,breaklinks,colorlinks,citecolor=cvprblue]{hyperref}
\usepackage{algorithm, algorithmic}
\usepackage{multirow}
\usepackage{bbding}

\title{Open-Vocabulary SAM3D: Towards Training-free Open-Vocabulary \\ 3D Scene Understanding}

\author{%
    Hanchen Tai$^1$\footnotemark[1]
  ~~  Qingdong He$^2$\thanks{Equal contributions.}
  ~~ Jiangning Zhang$^2$\thanks{Project leader.}
  ~~ Yijie Qian$^1$ \\
  ~~ Zhenyu Zhang$^3$
  ~~ Xiaobin Hu $^2$
  ~~ Xiangtai Li$^4$
  ~~ Yabiao Wang$^2$
  ~~ Yong Liu$^1$ \\
  \normalsize $^1$Zhejiang University ~~ $^2$Youtu Lab, Tencent~~ $^3$Nanjing University ~~ $^4$Nanyang Technological University\\
  \url{https://hithqd.github.io/projects/OV-SAM3D/}
}

\begin{document}
\maketitle
\begin{abstract}
Open-vocabulary 3D scene understanding presents a significant challenge in the field. 
Recent works have sought to transfer knowledge embedded in vision-language models from 2D to 3D domains. 
However, these approaches often require prior knowledge from specific 3D scene datasets, limiting their applicability in open-world scenarios. 
The Segment Anything Model (SAM) has demonstrated remarkable zero-shot segmentation capabilities, prompting us to investigate its potential for comprehending 3D scenes \textbf{without} training. 
In this paper, we introduce OV-SAM3D, a training-free method that contains a universal framework for understanding open-vocabulary 3D scenes. 
This framework is designed to perform understanding tasks for any 3D scene without requiring prior knowledge of the scene.
Specifically, our method is composed of two key sub-modules: 
First, we initiate the process by generating superpoints as the initial 3D prompts and refine these prompts using segment masks derived from SAM. 
%
Moreover, we then integrate a specially designed overlapping score table with open tags from the Recognize Anything Model (RAM) to produce final 3D instances with open-world labels.
%
Empirical evaluations on the ScanNet200 and nuScenes datasets demonstrate that our approach surpasses existing open-vocabulary methods in unknown open-world environments. 
Full codes and models will be made publicly available. 
\end{abstract}

\section{Introduction} 
\label{section:intro}
Recently, research on foundation models has been booming~\cite{devlin2018bert,brown2020language,raffel2020exploring,radford2021learning,ramesh2022hierarchical,fedus2022switch,achiam2023gpt,wu2024towards}, and they have shown emergent capabilities of complex reasoning, reasoning with knowledge, and out-of-distribution robustness~\cite{wei2022emergent}. 
Foundation models have profoundly influenced various fields beyond NLP~\cite{moor2023foundation,sam,tu2024towards}, particularly with the advent of the Segment Anything Model (SAM)~\cite{sam}, which has marked a significant breakthrough in the computer vision community.
The zero-shot segmentation performance of SAM is remarkable, prompting some efforts to extend its powerful generalization capabilities to other downstream tasks~\cite{ren2024grounded, ma2024segment, chen2023sam, yu2023inpaint}. One major transferable domain for such advancements is 3D scene segmentation and its further understanding~\cite{yang2023sam3d, xu2023sampro3d, yin2023sai3d, takmaz2023openmask3d}.

\begin{figure}
    \centering
    \includegraphics[width=1.0\linewidth]{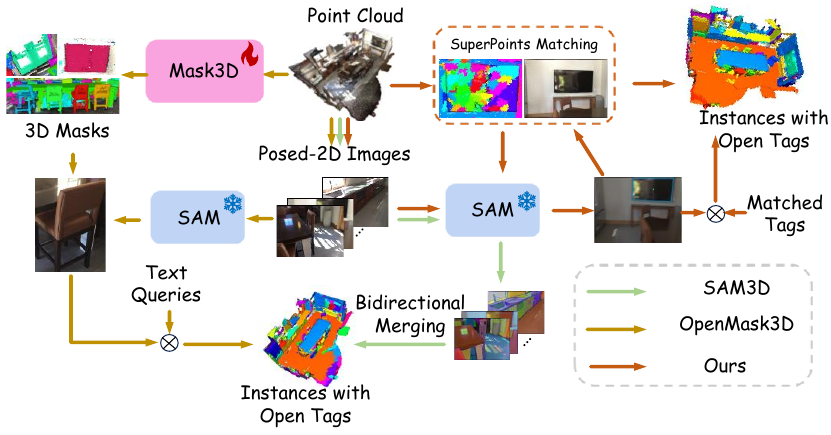}
    \caption{\textbf{Paradigm comparison with contemporary methods}. 
    \textbf{(a)}: SAM3D~\cite{yang2023sam3d} leverages SAM to achieve zero-shot 3D segmentation but only generates the class-agnostic segmentation results.
    \textbf{(b)}: OpenMask3D~\cite{takmaz2023openmask3d} implements open-vocabulary scene understanding but requires training a 3D proposal network under supervision.
    \textbf{(c)}: Our OV-SAM3D model effectively transfers the extensive knowledge embedded in the SAM to the 3D domain and enhances instance segmentation and recognition without requiring additional training. This capability enables open-vocabulary 3D scene understanding across various environments.
    }
    \label{fig1}
    \vspace{-0.2cm}
\end{figure}
Given the difficulties and costs associated with collecting 3D data, transferring knowledge from foundation models in the 2D domain to the 3D domain presents a viable solution for achieving zero-shot learning in 3D scene understanding.
Existing studies, such as SAM3D~\cite{yang2023sam3d}, SAMPro3D~\cite{xu2023sampro3d} and SAI3D~\cite{yin2023sai3d}, demonstrate effective 3D instance segmentation. However, these methods are constrained by the fact that SAM focuses solely on segmentation without incorporating recognition, thereby falling short of achieving full 3D scene understanding.
On the basis of these zero-shot segmentation methods, advancing open-vocabulary 3D scene understanding requires the model to locate and recognize objects within a 3D scene through text guidance, regardless of whether the objects have been seen before. Related works, such as OpenScene~\cite{peng2023openscene}, OpenMask3D~\cite{takmaz2023openmask3d} and Open3DIS~\cite{nguyen2024open3dis}, can achieve open-vocabulary 3D scene understanding when trained on datasets containing similar scenes. However, when the domain gap is substantial, nearly all of these methods fail, hindering the achievement of open-vocabulary 3D scene understanding in open-world scenarios. To address the challenge of understanding any 3D scene in an open-world, we propose a training-free framework that efficiently performs 3D scene understanding tasks without being constrained by domain gaps.

In this work, we introduce OV-SAM3D, a universal open-vocabulary 3D scene understanding framework capable of performing understanding tasks on any given scene without prior knowledge about the scene. Inspired by the classic graph-based segmentation algorithm~\cite{felzenszwalb2004efficient}, our OV-SAM3D introduces superpoints as initial 3D prompts, which are subsequently merged into rough 3D masks. Then we calculate the designed overlapping scores through the back-projected segmentation results of SAM~\cite{sam} for further refinement. By leveraging RAM's recognition capabilities in posed images of the 3D scene and utilizing ChatGPT for screening, OV-SAM3D can generate open-world image tags matched with the initial 3D prompts using vision-language models like CLIP~\cite{radford2021learning}. Based on the overlapping scores and matched tags, OV-SAM3D refines the rough 3D masks to produce final labeled 3D instances. In contrast to existing methods, our approach not only overcomes the limitations of class-agnostic masks in transferring SAM knowledge from 2D to 3D but also achieves superior results in open-vocabulary 3D scene understanding(Fig.~\ref{fig1}). As a training-free 3D scene understanding algorithm, OV-SAM3D can comprehend arbitrary 3D scenes without prior knowledge, such as the categories of objects present and the consistent features of the domain. 

To compare with existing methods, we conduct experiments on the well-studied ScanNet200 dataset~\cite{rozenberszki2022language}, but the closed-set environment goes against our original intention of understanding any 3D scene in an unknown open-world. Consequently, we also performed experiments on the outdoor nuScenes dataset~\cite{caesar2020nuscenes}, where few existing algorithms have been applied. The results show that OV-SAM3D is superior to existing state-of-the-art methods in an open-world context.

\begin{figure*}[!t]
    \centering
    \includegraphics[width=1\linewidth]{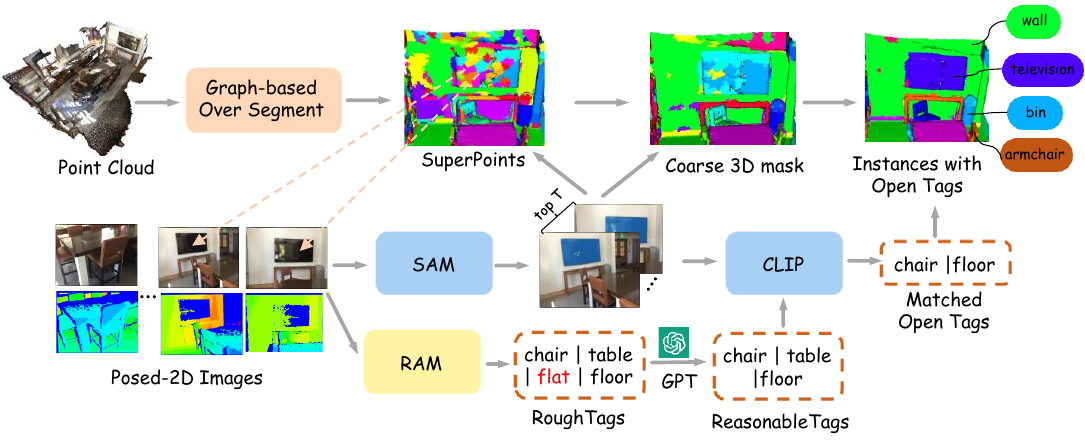}
    \caption{\textbf{Overview of our OV-SAM3D} that consists of two sub-modules: 
    1) SAM-centric Coarse Mask Generation first adapts a graph-based over-segmentation method to generate superpoints and selects some as initial 3D prompts to guide SAM. Then we revise the initial 3D prompts through the masks of SAM and create an overlapping score table.
    2) Open Tags Guided Coarse Mask Merging combines the overlapping score table and reasonable open tags recognized by RAM, thus we can achieve the open-vocabulary 3D scene understanding task to get 3D instances with labels.
    }
    \label{fig2}
\end{figure*}

Overall, our contributions are three-fold:
\begin{itemize}
    \item We present OV-SAM3D, a universal framework for open-vocabulary 3D scene understanding, capable of understanding any 3D scenes without prior knowledge.
    \item Our approach innovatively employs superpoints to generate initial coarse 3D masks via back-projection from the Segment Anything Model (SAM), which are subsequently refined and merged using filtered open tags and a specially designed overlapping score algorithm.
    \item Extensive experimental evaluations on the ScanNet200 and nuScenes datasets demonstrate the superior performance of OV-SAM3D, showcasing its effectiveness in unknown open-world scenarios.
\end{itemize}

\section{Related Work} 
\label{section:related}

\noindent
\textbf{Multimodal Foundation Models.}
The foundational models have gradually solidified their critical importance in the broader domain of artificial intelligence~\cite{sarlin2020superglue,sakaguchi2021winogrande, fedus2022switch}. In particular, multimodal foundational models, which integrate data from various modalities, have emerged as a prominent area of research~\cite{radford2021learning,feng2022promptdet,fang2023eva,yuan2024open}. CLIP~\cite{radford2021learning}conducts contrastive learning on large datasets, enabling the model to classify various images effectively. Based on this, MaskCLIP~\cite{zhou2022extract} enhances the image-text classification capabilities by employing feature upsampling, extending functionality to coarse image segmentation. Tag2Text~\cite{huang2023tag2text} introduces a foundational image tagging model that generates a broader range of tag categories and embeds these image tags into the visual-language model to guide downstream tasks. RAM~\cite{zhang2023recognize} further innovates by combining tagging and captioning tasks, thereby expanding the recognition capabilities of the foundational image tagging model. As a landmark foundation model in the vision community, SAM~\cite{sam} can generate high-quality masks through point prompts, box prompts, and text prompts. Through deep integration of SAM and CLIP, OV-SAM~\cite{yuan2024open} can generate high-quality semantic segmentation masks for any image. Our research attempts to effectively combine the recognition capability of RAM, the localization strengths of SAM, and the classification power of CLIP to accomplish comprehensive 3D scene understanding tasks.

\noindent
\textbf{Zero-Shot 3D Scene Segmentation.}
The impressive performance of SAM in zero-shot image segmentation tasks has greatly advanced the development of foundation models in computer vision. Its derivative models can be applied to challenging fields of segmentation such as medical images and remote sensing images and even extended to 3D point clouds for zero-shot 3D segmentation~\cite{chen2023sam, cheng2023segment,ren2024grounded,ma2024segment,cen2023segment,he2024pointseg}. Recent research increasingly relies on SAM to implement foundational model functions in the 3D domain due to the high cost of collecting extensive 3D point cloud data. SAM3D~\cite{yang2023sam3d} projects multiple segmentation masks generated by SAM's automatic segmentation onto a 3D scene, where adjacent point clouds are iteratively merged to produce a comprehensive 3D mask that encompasses the entire scene. Following the preliminary attempts with SAM3D, SAMPro3D~\cite{xu2023sampro3d} refines the segmentation process by sampling points from the 3D scene and projecting them into the posed images as point prompts to aid SAM in achieving more accurate segmentation results. SAI3D~\cite{yin2023sai3d} introduces over-segmented results, which are then merged using affinity to generate the final 3D segmentation masks. However, these zero-shot 3D scene segmentation methods inherit the class-agnostic nature from SAM, making them unable to recognize specific segmented instances. In contrast to previous methods, our research does not rely on generating class-agnostic 3D masks followed by combining them with models like CLIP~\cite{radford2021learning} or GroundingDINO~\cite{liu2023grounding} to obtain labels. Instead, we embed relevant semantic information directly into the mask generation process, enabling us to better address the challenges of understanding any 3D scene.

\noindent
\textbf{Open-Vocabulary 3D Scene Understanding.}
Given any ambient text queries, open-vocabulary 3D scene understanding tasks~\cite{ding2023pla,he2024unim} aims to locate and identify matching instances, regardless of whether the model has previously encountered these instances. OpenScene~\cite{peng2023openscene} leverages the frozen 2D scene understanding model~\cite{ghiasi2022scaling,li2022language} to train an aligned 3D encoder for extraction of dense features from point clouds in a 3D scene, and then ensembles the 2D\&3D features o closely match arbitrary text queries. OpenIns3D~\cite{huang2023openins3d} first employs a trained 3D network to generate class-agnostic mask proposals, which are then projected onto synthetic scene-level images~\cite{bakr2022look,kundu2020virtual}. With the 2D scene understanding model detecting the position and category of objects in these scene-level images, OpenIns3D matches the projected mask proposals to achieve open-vocabulary 3D scene understanding. OpenMask3D~\cite{takmaz2023openmask3d} also utilizes Mask3D~\cite{schult2023mask3d} to generate class-agnostic masks for 3D scenes, which are projected onto the posed images as point prompts. The method then extracts average CLIP embeddings from multi-scale image crops derived from SAM's segmentation results, matching these embeddings with the query texts to segment 3D instances. Compared to previous studies, the significant improvement of our method is that we abandon training-required 3D encoders, which limit the generalization ability of models, and explore the collaborative effects of foundation models to achieve a universal open-vocabulary 3D scene understanding algorithm.

\section{Methodology} \label{section:method}
Fig.~\ref{fig2} shows the detailed framework of our OV-SAM3D model. Given any scene and its associated set of posed RGB-D images, we draw inspiration from recent works~\cite{yang2023sam3d, yin2023sai3d} to generate approximately over-segmented superpoints. The core of OV-SAM3D lies in merging these superpoints into accurate 3D instances and obtaining corresponding labels throughout the process. Firstly, we select high-quality superpoints as the initial 3D prompts, which are then projected onto posed images as point prompts to guide SAM in fine segmentation. The segmented mask is subsequently projected back into the 3D scene to refine the initial 3D prompts into coarse masks, from which overlapping scores are extracted. Concurrently, we utilize RAM~\cite{zhang2023recognize} to identify open tags from the corresponding frames and employ ChatGPT to filter out unreasonable tags. By combining overlapping scores with matched open tags using CLIP, we obtain the final 3D instances and their labels.

\subsection{SAM-centric Coarse Mask Generation}
\label{Coarse Masks}
\paragraph{Generate SAM Masks Guided by 3D Prompts.}
Directly processing unstructured 3D point clouds in 3D scenes significantly impacts computational efficiency, while grouping 3D points with similar geometric properties into continuous regions as superpoints remarkably accelerates the processing of OV-SAM3D. Drawing on the graph based image segmentation algorithm~\cite{felzenszwalb2004efficient}, 
we generate superpoints $\{S_m\in \mathbb{R}^{P_m\times3}\}_{m=1}^{M}$ and select high-quality superpoints as the initial 3D prompts $\{Q_n\}_{n=1}^{N}\subseteq S$ based on $P_m$, which denotes the number of points in superpoint $S_m$. We then project the 3D prompts onto the top $T$ views with the lowest occlusion levels, assessed by the number of visible points in the depth pictures, and sample the projected pixel points as point prompts to guide SAM in fine segmentation.

\paragraph{Revise 3D Prompts via Back-projection.}
Considering the initial 3D prompts as the origins of the final 3D instance, we design an overlapping score table $F\in \mathbb{R}^{M\times N}$ to determine the association of all superpoints with their respective origins. We then back-project the $n$-th 2D mask onto the $t$-th view as $BP_{nt}\in \mathbb{R}^{P_{nt}\times3}, n \leq N, t \leq T$, to revise the randomly segmented 3D mask into a coarse 3D mask, where $P_{nt}$ denotes the number of points included in the 3D back-projection.

    \begin{algorithm}[H] 
      \caption{Calculate overlapping score table} 
      \label{alg1} 
      \renewcommand{\algorithmicrequire}{\textbf{Input:}}
      \renewcommand{\algorithmicensure}{\textbf{Output:}}
      \begin{algorithmic}
        \REQUIRE 
        superpoints $\{S_m\in \mathbb{R}^{P_m\times3}\}_{m=1}^{M}$, \\
        Initial 3D prompts $\{Q_n\}_{n=1}^{N}$, \\
        3D back-projection $\{BP_{nt}\in \mathbb{R}^{P_{nt}\times3}\}_{n,t=1}^{N,T}$
        \ENSURE
        Overlapping score table $F\in \mathbb{R}^{M\times N}$ 
        \STATE $n \gets 1$, $F \gets \mathbf{0}$, threshold $\gets \theta$
        \WHILE{$n < N$} 
        \STATE $t \gets 1$
        \WHILE{$t < T$}
        \STATE $m \gets 1$
        \WHILE{$m < M$}
        \STATE overlapping radio $\gets \frac{S_{m}\cap BP_{nt}}{S_{m}}$   
        \IF{overlapping radio $> \theta$} 
        \STATE $F_{(m-1)(n-1)} \gets F_{(m-1)(n-1)} + 1$ 
        \ENDIF
        \STATE $m \gets m + 1$ 
        \ENDWHILE
        \STATE $t \gets t + 1$ 
        \ENDWHILE
        \STATE $n \gets n + 1$
        \ENDWHILE  
      \end{algorithmic} 
    \end{algorithm}
    
\begin{figure}[H]
\centering
\includegraphics[width=1.0\linewidth]{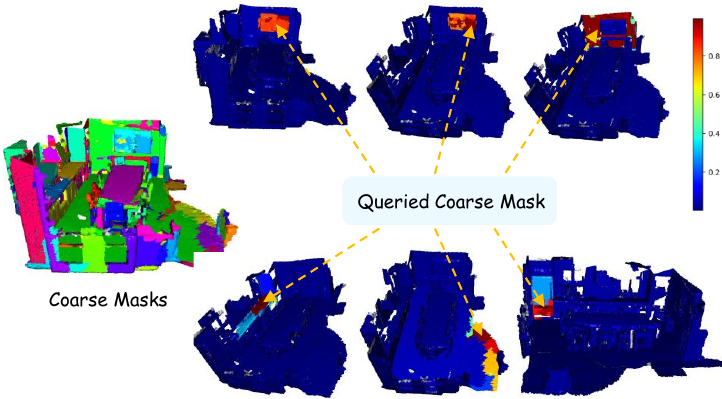}
\caption{\textbf{Visualization of overlapping scores}. We calculate the overlapping scores between queried coarse mask with other all other coarse masks. If the score is higher than a certain threshold, it indicates that these two coarse masks belong to the same instance. Here, we visualize some overlapping scores from \textcolor{red}{high} to \textcolor{blue}{low}.}
\label{fig_vis}
\end{figure}

The basis of this revision process and subsequent merging procedure is the designed overlapping score table $F$, generated as outlined in Algorithm~\ref{alg1}. Firstly, we calculate the overlap region of all superpoints with the back-projection in the same 3D scene. Then, we update the score at the corresponding position $F_{ij}$, based on whether the overlap region of $\{S_m\cap Q_n\}$ meets a certain threshold $\theta$. After merging all superpoints into their respective 3D prompts, we obtain the overlapping score table $F$, as visualized in Fig.~\ref{fig_vis}. In this table, along the row axis, the maximum index in each row $F_m^{1\times N}$ indicates the 3D prompt to which superpoint $S_m$ should belong, allowing us to merge all superpoints into the corresponding 3D prompts, resulting in coarse 3D masks $R_c\in \mathbb{R}^{P\times N}$, where $P$ denotes the total number of points in the scene. Along the column axis, each column $F_n^{M\times1}$ represents the overlap distribution between each coarse 3D mask and all superpoints, which we define as the overlapping score of that coarse 3D mask.





\subsection{Open Tags Guided Coarse Mask Merging}
\label{Merge Masks}
\paragraph{Merge Coarse Masks.}
To generate final high-quality 3D instance masks, it is necessary to remerge the initial coarse 3D masks due to the uncertainty associated with the initial 3D prompts in any given scene. This remerging process is determined by whether the total score $TS_{n}\in \mathbb{R}^{M\times 1}$, representing the overlap between coarse 3D masks, exceeds a certain threshold. This threshold is derived from the overlap score itself and dynamic parameter selected by the user.
\begin{equation}
    TS_{n}=F_{n}^{T}\times F>\max(1,|F_{n}|/\tau), 
    \label{eq1}
\end{equation}
where $TS_{n}$ is the similarity score of $n$-th coarse 3D mask with other masks. $\tau$ is a parameter that can be modified by user-interaction. 
Note that OV-SAM3D provides several user-adjustable hyperparameters, with only their ranges constrained. This allows users to better adapt the model for any 3D scene in an open-world environment. 

\begin{figure}
\includegraphics[width=1\linewidth]{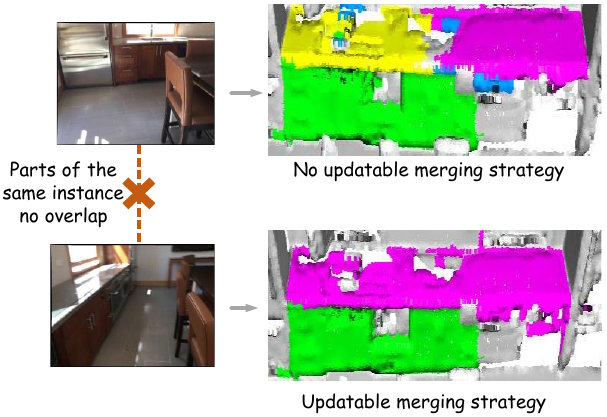}
\caption{\textbf{What issues can updatable merging strategy solve ?} When a large instance is separated in different views, their overlap region may be minimal and cannot be directly merged as shown on the top. Through our updatable merging strategy, the progressive mask can gradually enlarge and meet the merging conditions.}
\label{fig_video}
\vspace{-0.2cm}
\end{figure}

Limited by the narrow perspective of posed images, some large instances may not have overlapping regions across different views, making it impossible to merge them directly based on overlapping scores. As shown in Fig.~\ref{fig_video}, we employ an updatable merging strategy that continuously updates the coarse 3D mask to reflect the merged result while recording the composition of the merged coarse 3D masks throughout the process. This approach ultimately enables the generation of 3D instances along with detailed composition records.

\begin{figure}
    \centering
    \includegraphics[width=1\linewidth]{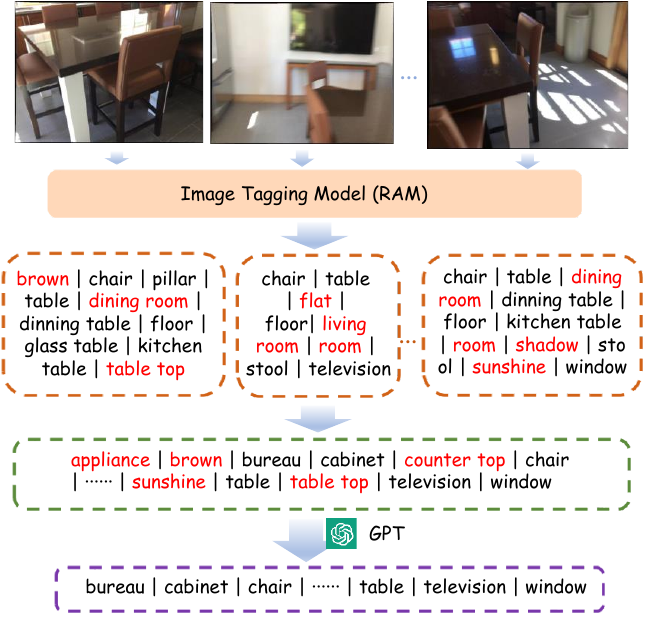}
    \caption{Generation of open instance tags. We show how to generate the open instance tags through RAM~\cite{zhang2023recognize} and ChatGPT. The red \textcolor{red}{tags} from RAM are not instance labels that need to be filtered by ChatGPT.}
    \label{ram}
\vspace{-0.2cm}
\end{figure}

\paragraph{Recognize Open Instance Tags.}
OV-SAM3D implements a universal open 3D scene understanding framework with the strict requirement that the model cannot acquire any prior knowledge of the 3D scenes it is intended to understand. Without prior knowledge of potential objects in a 3D scene, we introduce RAM to perform preliminary recognition on the corresponding posed images of the scene, as shown in Fig.~\ref{ram}. RAM provides a rough identification of which instances or labels are present. However, the open tag library of RAM includes not only instance tags but also tags representing colors, material properties, room types, and other instance-independent tags that are not conducive to 3D scene understanding tasks. To address this, we employ ChatGPT as a filter following RAM to select only the instance tags recognized by RAM, thereby generating the final open instance tags.

\paragraph{Match Open Instance Tags.}
After merging coarse masks into 3D instances and recognizing open instance tags, we match the tags with the 3D instance segmentation masks through CLIP to achieve the open-vocabulary 3D scene understanding task. To maintain matching accuracy without increasing algorithmic complexity, we avoid reprojecting the 3D masks onto posed images. Instead, we extract image crop features from existing SAM-segmented 2D masks using CLIP's image encoder and match the CLIP text features $FT \in \mathbb{R}^{C \times 768}$ extracted from open tags with the image crop features $\{FC_k \in \mathbb{R}^{I_k \times 768}\}_{k=1}^{K}$, where $I_k$ represents the number of coarse masks in the $k$-th instance according to the composition records mentioned before:
\begin{equation}
     CS_{ki}=\frac{FC_{ki}\cdot FT}{|FC_{ki}||FT|},
\end{equation}
where $CS_{ki}$ represents the cosine similarity between the $i$-th course mask in the $k$-th instance and each of the $C$ text queries. For each coarse mask, we can choose the tag with the highest similarity as the label $\{Label_i\}_{i=1}^{I_k}$, but the matching result cannot be directly used as the label of the final 3D instance. We list the coarse 3D masks within each 3D instance based on the composition records and align the open tags $Label_i$ with the highest feature similarity $CS_{ki}$ as the final label $Label_k$ for the corresponding 3D segmentation instance.
Subsequent ablation study in Tab.~\ref{tab:3} demonstrates the effectiveness of the strategy of selecting the open-label with the highest feature similarity as the final 3D instance label.
\section{Experiments} \label{section:exp}
To quantitatively evaluate our method, we compare OV-SAM3D with some leading open-vocabulary 3D scene understanding methods and 3D instance segmentation methods suitable for 3D scene understanding tasks on the well-studied dataset ScanNet200~\cite{rozenberszki2022language} and the outdoor dataset nuScenes~\cite{caesar2020nuscenes}. Furthermore, we provide ablation studies for OV-SAM3D, explaining the improvement of each module in our framework and the impact of different hyper-parameters. In addition, we display some qualitative results of our method in any 3D scene, more qualitative results can be seen in the Appendix.

\subsection{Experimental Setup}
\paragraph{Datasets.}
ScanNet200~\cite{rozenberszki2022language} is an RGB-D dataset commonly utilized for evaluating open-vocabulary 3D scene understanding tasks, which expands the original ScanNet dataset~\cite{dai2017scannet} by labeling a more extensive set of 200 instance categories. ScanNet200 consists of 1,513 indoor scenes divided into 1,201 training scenes and 312 validation scenes, which contains surface mesh point clouds, instance-level semantic annotations, and semantic labels for each point. In contrast, nuScenes~\cite{caesar2020nuscenes} is an outdoor dataset comprising 850 training scenes, 150 validation scenes, and 150 testing scenes. It provides various sensor data for multiple 3D perception tasks, such as 3D object detection, instance segmentation, panoramic segmentation, and object tracking. Given that existing open-vocabulary 3D scene understanding algorithms are rarely applied to outdoor datasets, we conduct supplementary experiments on nuScenes to evaluate the effectiveness of OV-SAM3D across diverse scenes.
\paragraph{Evaluation metrics.}
Following the setting of the baseline~\cite{takmaz2023openmask3d}, we choose the average precision score to evaluate the quantitative results. We respectively calculate the score at the average mask overlap thresholds between 50\% and 95\% along 5\% steps as AP and at the mask overlap threshold of 25\%, 50\% as $\mathrm{AP_{25}}$, $\mathrm{AP_{50}}$. For ScanNet200 dataset, we calculate AP under different category groups~\cite{rozenberszki2022language}, namely $\mathrm{AP_{head}}$, $\mathrm{AP_{com}}$ and $\mathrm{AP_{tail}}$. For the nuScenes dataset, we calculate the average precision for the three main categories of car, truck, and bus as $\mathrm{AP_{car}}$, $\mathrm{AP_{truck}}$ and $\mathrm{AP_{bus}}$.
\paragraph{Implementation details.}
\label{details}
As an open-vocabulary 3D scene understanding framework that does not require prior knowledge of datasets, OV-SAM3D sets some interactive hyper-parameters for arbitrary 3D scenes. To enable comparison with other existing algorithms on the close-dataset, we fix these hyper-parameters, which significantly affects the performance of our algorithm. For ScanNet200, we select 200 high-quality superpoints as initial prompts, set the threshold $\tau$ in Eq.~\ref{eq1} to 0.45 and sample RGB-D frames at a frequency of 10. For nuScenes, we select 100 high-quality superpoints and keep the threshold $\tau$. In addition, more details about converting nuScenes data formats are provided in the Appendix. Following these settings, the task of performing a comprehensive scene understanding, which includes both segmentation and recognition, requires an average of 4-6 minutes on a ScanNet scene and 30-60 seconds on a nuScenes scene when executed on a single GPU.

\subsection{Quantitative 3D scene understanding evaluation}
To ensure consistency with existing open-vocabulary 3D scene understanding methods, we conducted quantitative experiments on the well-studied ScanNet200 dataset. However, it is important to note that some methods had prior knowledge of ScanNet200, which diverges from our original goal of understanding any 3D scene in unknown environments. Therefore, we extended our experiments to the nuScenes dataset to further demonstrate the superiority of our method in understanding any 3D scene.

\paragraph{Comparison on relatively open understanding tasks.}
The numerical results on ScanNet200 are presented in Tab.~\ref{tab:1}, where we compare OV-SAM3D with two types of algorithms: those requiring pre-training on ScanNet200 and those adaptable to open-vocabulary 3D scene understanding tasks without such training. In scenarios where pre-training is permitted, our advantage of understanding any open-world 3D scene seems difficult to demonstrate, but even so, we reflect a certain level of competitiveness especially in $\mathrm{AP_{25}}$ and $\mathrm{AP_{tail}}$. In order to further highlight our research objectives, we compare with the state-of-the-art method OpenMask3D on the outdoor nuScenes dataset that is blank for all algorithms.

\begin{table*}[!t]
  \centering  
  \resizebox{1.0\linewidth}{!}{
    \begin{tabular}{ccccccccc}
        \toprule
        Method & 3D Proposal & Semantic & $\mathrm{AP}$ & $\mathrm{AP_{50}}$ & $\mathrm{AP_{25}}$ & $\mathrm{AP_{head}}$ & $\mathrm{AP_{com}}$ & $\mathrm{AP_{tail}}$ \\
        \midrule
        \textit{Training-based} \\
        OpenScene~\cite{peng2023openscene} & 3D Distill & OpenSeg~\cite{ghiasi2022scaling} & 4.8 & 6.2 & 7.2 & 10.6 & 2.6 & 0.7 \\
        OpenScene~\cite{peng2023openscene} & 2D/3D Ensemble & OpenSeg~\cite{ghiasi2022scaling} & 5.3 & 6.7 & 8.1 & 11.0 & 3.2 & 1.1 \\
        OpenScene~\cite{peng2023openscene} & Mask3D~\cite{schult2023mask3d} & OpenSeg~\cite{ghiasi2022scaling} & \underline{11.7} & \underline{15.2} & 17.8 & \underline{13.4} & \underline{11.6} & 9.9 \\
        OpenMask3D~\cite{takmaz2023openmask3d} & Mask3D~\cite{schult2023mask3d} & CLIP~\cite{radford2021learning} & \textbf{15.4} & \textbf{19.9} & \textbf{23.1} & \textbf{17.1} & \textbf{14.1} & \textbf{14.9} \\
        \midrule
        \textit{Training-free} \\
        SAM3D~\cite{yang2023sam3d} & None & OpenSeg~\cite{ghiasi2022scaling} & 7.4 & 11.2 & 16.2 & 6.7 & 8.0 & 7.6 \\
        SAI3D~\cite{yin2023sai3d} & None & OpenSeg~\cite{ghiasi2022scaling} & 9.6 & 14.7 & 19.0 & 9.2 & 10.5 & 9.1 \\
        Ours & None & CLIP~\cite{radford2021learning} & 9.0 & 13.6 & \underline{19.4} & 9.1 & 7.5 & \underline{10.8} \\
        \bottomrule
    \end{tabular}
    }   
  \caption{3D scene understanding results on the ScanNet200 validation set. In the table, the best results are in \textbf{bold} while the second best results are \underline{underscored}. We compare against both methods with training and methods without training, and report different evaluation indicators, in which $\mathrm{AP_{tail}}$ can evaluate the performance of models on the long-tail distribution.} 
  \label{tab:1}
  \vspace{0.2cm}
\end{table*}

\paragraph{Comparison on more open understanding tasks.}
As a high-quality dataset for autonomous driving, the nuScenes dataset is seldom utilized in the task of open-vocabulary 3D scene understanding. Therefore, we supplement some experiment on this less-explored dataset to compare the capabilities of models in the open world. Most advanced methods that require training~\cite{nguyen2024open3dis,lu2023ovir} have been trained exclusively on indoor datasets, and directly applying or finetuning them on the nuScenes dataset would undermine the validity of the experiment. Therefore, we select OpenMask3D~\cite{takmaz2023openmask3d}, which provides a publicly available version capable of running on arbitrary scenes. For methods that do not require training, the typical approach involves a two-stage process: first, generating class-agnostic 3D masks, followed by integration with existing 3D scene understanding frameworks. In our experiments, we consistently utilized OpenMask3D for this phase. As shown in Tab.~\ref{tab:2}, these state-of-the-art methods are clearly not competent on the outdoor nuScenes dataset, while our OV-SAM3D still maintains impressive understanding performance.

\begin{table*}[ht]
  \centering
  \resizebox{1\linewidth}{!}{
    \begin{tabular}{cccccccccc}
      \toprule
        Method & Setting & $\mathrm{AP}$ & $\mathrm{AP_{50}}$ & $\mathrm{AP_{25}}$ & $\mathrm{AP_{car}}$ & $\mathrm{AP_{truck}}$ & $\mathrm{AP_{bus}}$\\
        \midrule
        OpenMask3D~\cite{takmaz2023openmask3d} & Training for arbitrary scenes & 0.5 & 1.4 & 5.2 & 0.3 & 0.2 & 0.6\\
        \midrule
        SAMPro3D~\cite{xu2023sampro3d} + OpenMask3D & Training-free & 0.8 & 1.1 & 5.2 & 0.6 & 0.0 & 0.2\\
        SAI3D~\cite{yin2023sai3d} + OpenMask3D & Training-free & 2.8 & 5.4 & 7.2 & 1.9 & 2.6 & 22.4\\        
        Ours & Training-free & \textbf{8.9} & \textbf{16.0} & \textbf{29.1} & \textbf{13.6} & \textbf{16.2} & \textbf{39.8}\\
        \bottomrule
      \end{tabular}
  }
  \caption{3D scene understanding results on the nuScenes validation set. Following the setup of SAI3D, we uniformly use OpenMask3D as the model for matching class-agnostic masks with text features. For the complete OpenMask3D, we choose its publicly available model that adapts to any 3D scene.}
  \label{tab:2}
  \vspace{0.2cm}
\end{table*}

\subsection{Qualitative results}
\begin{figure}[H]
    \centering
    \includegraphics[width=1\linewidth]{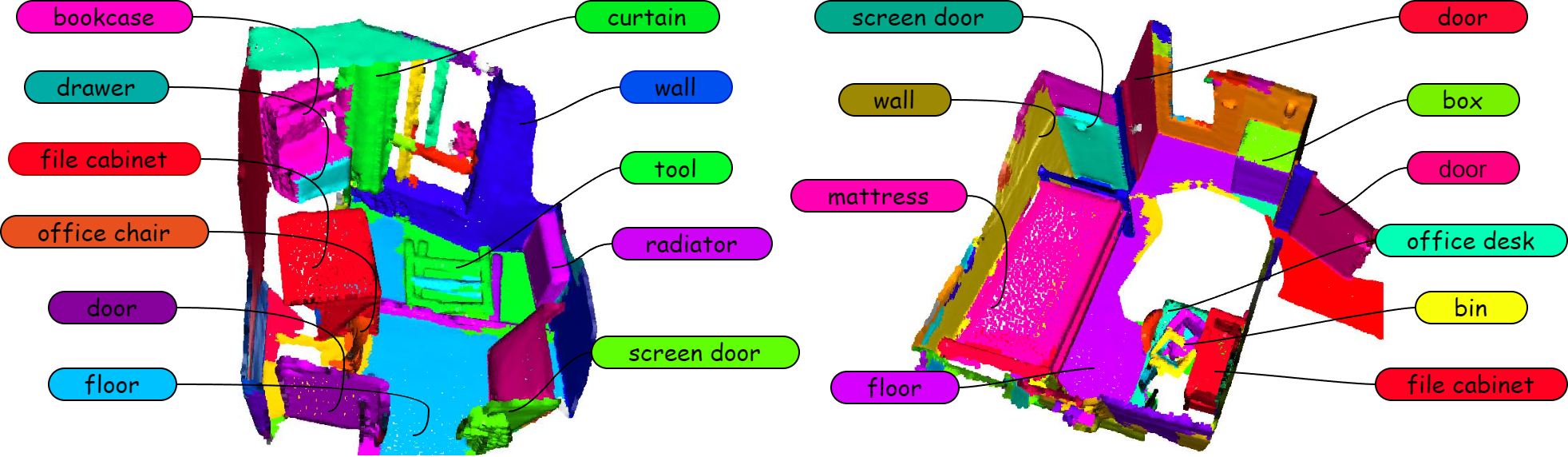}
    \caption{\textbf{Qualitative results of our method on open-world}. Even if the instance categories present in the scene are not known in advance, OV-SAM3D can still utilize embedded open-tags to achieve 3D scene understanding tasks.}
    \label{fig:6}
\end{figure}

\begin{figure}[H]
    \centering
    \includegraphics[width=1\linewidth]{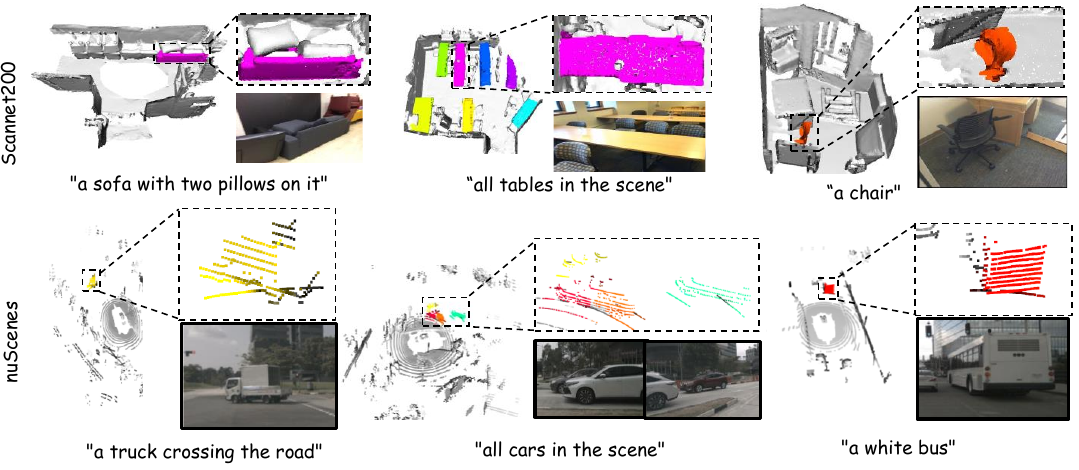}
    \caption{Qualitative results of our method on the ScanNet200 datasets and the nuScenes datasets with various text queries.}
    \label{fig:7}
    \vspace{0.2cm}
\end{figure}

Here we show some qualitative results from our OV-SAM3D. In Fig.~\ref{fig:6}, our approach achieves the 3D scene understanding in an open-world environment, automatically recognizing various instances in the scene without requiring a text query, due to its embedded open tags. In Fig.~\ref{fig:7}, we present the text-driven results of OV-SAM3D on the ScanNet200 dataset and nuScenes dataset. Not limited to simple category text queries, OV-SAM3D demonstrates an enhanced ability to comprehend the entire 3D scene and detect objects, even with complex language cues.

\subsection{Ablation study}
In this section, we discuss the influence of different components of OV-SAM3D on scene understanding, and the effectiveness of each module design can be proved from Tab.~\ref{tab:3}. As mentioned before, OV-SAM3D retains some interactive hyper-parameters, and in Tab.\ref{tab:4}, we present the results of different hyper-parameters.

\paragraph{Study on different components.}
The core designed modules of OV-SAM3D include backprojecting under SAM supervision, overlapping feature similarity remerging of coarse 3D masks, and matching strategies for 3D instances with open tags: the highest feature similarity or the highest matching number.
\begin{table}[ht]
  \centering
  \resizebox{1\linewidth}{!}{
    \begin{tabular}{cccccccccc}
      \toprule
        \multirow{2}{*}{Backproject} & \multirow{2}{*}{Remerge} & \multicolumn{2}{c}{Match Tags} & \multirow{2}{*}{$\mathrm{AP}$} & \multirow{2}{*}{$\mathrm{AP_{50}}$} & \multirow{2}{*}{$\mathrm{AP_{25}}$} & \multirow{2}{*}{$\mathrm{AP_{head}}$} & \multirow{2}{*}{$\mathrm{AP_{com}}$} & \multirow{2}{*}{$\mathrm{AP_{tail}}$} \\
        \cmidrule{3-4}
         & & Score & Number \\
        \midrule
        \XSolidBrush & \XSolidBrush & \Checkmark & \XSolidBrush & 2.2 & 4.4 & 9.8 & 1.1 & 1.5 & 4.2 \\
        \Checkmark & \XSolidBrush & \Checkmark & \XSolidBrush & 6.5 & 11.4 & 17.8 & 5.2 & 5.3 & 9.6 \\        
        \Checkmark & \Checkmark & \Checkmark & \XSolidBrush & \textbf{9.0} & \textbf{13.6} & \textbf{19.4} & 9.1 & \textbf{7.5} & \textbf{10.8} \\         
        \Checkmark & \Checkmark & \XSolidBrush & \Checkmark & 8.8 & 13.5 & 19.3 & \textbf{9.2} & 7.4 & 9.9 \\          
        \bottomrule
      \end{tabular}
  }
  \caption{The ablation studies on OV-SAM3D designed modules. The experiments are conducted on ScanNet200 validation set and we fix hyper-parameters initial 3D prompts as 200, merging threshold $\tau$ as 0.45.}  
  \label{tab:3}  
\end{table}

\paragraph{Study on different hyper-parameters.}
The interactive hyper-parameters in OV-SAM3D are designed for users to understand any 3D scene. On large-scale datasets, such fixed hyperparameters are difficult to accommodate all scenes. In Tab.~\ref{tab:4}, we demonstrate the potential impact of different hyper-parameter selections.

\begin{table}[H]
  \centering
  \resizebox{1\linewidth}{!}{
    \begin{tabular}{cccccccc}
      \toprule
        Initial 3D Prompts & Merging threshold & $\mathrm{AP}$ & $\mathrm{AP_{50}}$ & $\mathrm{AP_{25}}$ & $\mathrm{AP_{head}}$ & $\mathrm{AP_{com}}$ & $\mathrm{AP_{tail}}$ \\
        \midrule
        150 & 0.50 & 7.8 & 12.9 & 18.6 & 8.2 & 6.0 & 9.4 \\        
        150 & 0.45 & 8.3 & 13.1 & 18.2 & 8.7 & 6.5 & 9.9 \\         
        200 & 0.45 & \textbf{9.0} & \textbf{13.6} & \textbf{19.4} & \textbf{9.1} & \textbf{7.5} & \textbf{10.8} \\
        200 & 0.50 & 9.0 & 13.6 & 18.8 & 9.1 & 7.4 & 10.6 \\
        \bottomrule
      \end{tabular}
  }
  \caption{Ablation studies on hyperparameters. The experiments are conducted on ScanNet200 validation set and we keep selecting tags with the highest feature similarity as instance labels.}  
  \label{tab:4}
\end{table}

\section{Conclusion and Future Works} \label{section:con}
We propose OV-SAM3D, a pioneering framework for open-vocabulary 3D scene understanding in open-world environments. Compared to existing algorithms, OV-SAM3D can understand arbitrary scenes without prior knowledge of the given environment. Through effectively exploring and leveraging the complementary strengths of foundational models, we have extended OV-SAM3D to a more challenging and meaningful open-world environment, a direction that has also been advocated in related foundational model research. Comparative experiments on the ScanNet200 and nuScenes datasets reveal that OV-SAM3D not only delivers competitive results on the well-studied ScanNet200 dataset but also significantly outperforms on the open nuScenes dataset, demonstrating its capability to understand any 3D scene in an unknown open-world. As a universal open-vocabulary 3D scene understanding model, OV-SAM3D inherits its comprehension capabilities from the foundational models, which is both its limitation and advantage. With the gradual development of foundation models, OV-SAM3D is expected to achieve further significant breakthroughs.


{
    \small
    \bibliographystyle{ieeenat_fullname}
    \bibliography{main}
}


\renewcommand\thefigure{A\arabic{figure}}
\renewcommand\thetable{A\arabic{table}}  
\renewcommand\theequation{A\arabic{equation}}
\setcounter{equation}{0}
\setcounter{table}{0}
\setcounter{figure}{0}
\appendix
\section*{Appendix}
\textbf{Overview}

The supplementary material presents the following sections to strengthen the main manuscript:

\begin{itemize}
\item[—] \textbf{Sec.A}~\ref{appendix:a} shows more experimental processing details for nuScenes dataset.
\item[—] \textbf{Sec.B}~\ref{appendix:b} shows more visualization results of Open Vocabulary SAM3D.
\end{itemize}

\section{Implementation Details on nuScenes} 
\label{appendix:a}
As a commonly used dataset for 3D segmentation and understanding, the preprocessing and evaluation procedures of the ScanNet200 dataset are already quite mature, and we follow the same setup as previous work.  Compared to the ScanNet200 dataset, which focuses on indoor scenes, the nuScenes dataset provides comprehensive autonomous driving data collected in various outdoor environments, resulting in different data representations and types. To achieve a unified evaluation framework, it is crucial to preprocess the nuScenes data, particularly its point clouds and instance annotations, to conform to the format and structure of ScanNet200. This process involves two main tasks: point cloud processing and instance assignment.
\subsection{Point Cloud Processing}
In the ScanNet200 dataset, point cloud data is typically stored in a mesh format, directly providing detailed surface information. However, the nuScenes dataset stores point clouds as raw coordinate data. To facilitate the generation of superpoints, we should convert the nuScenes point clouds into mesh representations first.
\paragraph{Study on different components.} The conversion from raw point clouds to meshes involves employing surface reconstruction techniques. Specifically, we use an Alpha-Shape-based surface reconstruction method, which consists of several steps: point cloud preprocessing, Alpha shape computation, mesh generation, and optimization. These steps are used to create a polygon mesh from the point cloud data, effectively restoring the surface structure. To balance noise reduction and detail preservation, we set the parameter $\alpha$ to 0.5, ultimately converting the nuScenes point cloud data into mesh representations compatible with the ScanNet200 data structure.
\paragraph{Superpoint Segmentation.} Superpoints are generated based on the mesh, ensuring that the parameters align with those used in the ScanNet200 dataset. Specifically, we select the top 100 superpoints as the initial 3D prompts based on the number of points in the cloud to maintain high relevance and detail.
\subsection{Instance Assignment}
The nuScenes dataset lacks explicit point cloud instance annotations, which are critical for various qualitative evaluations. To overcome this limitation, we employ the bounding box annotations provided by nuScenes to infer instance-level segmentation.
\paragraph{Point Cloud Segmentation.} The instance assignment process involves segmenting the point cloud based on bounding box annotations. We extract the point cloud fragments within each annotated box, effectively creating instance-level segments.
\paragraph{Bounding Box Extraction.} The instance assignment process entails segmenting the point cloud based on bounding box annotations. We isolate the point cloud fragments within each annotated bounding box, thereby creating preliminary instance fragments. To ensure each segment has sufficient point cloud density, we filter out annotation instances containing an insufficient number of points.
\paragraph{Label Merging. } The nuScenes dataset offers a comprehensive set of annotated box categories. To balance the number of instances and ensure sufficient point cloud density, we filtered and merged the annotation box categories, resulting in ten common category labels.

\section{More Qualitative Results} 
\label{appendix:b}
\begin{figure}[h]
    \centering
    \includegraphics[width=1\linewidth]{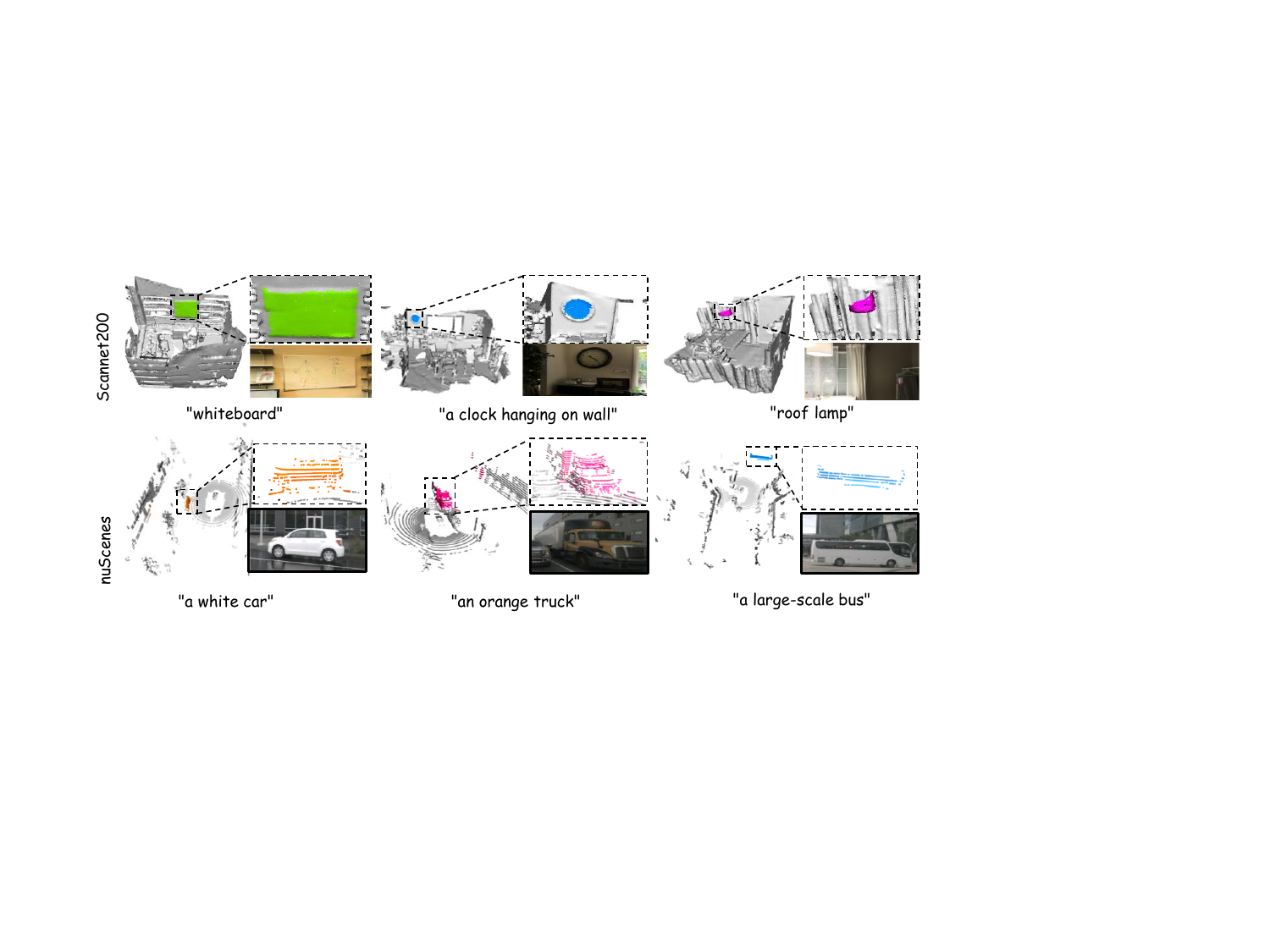}
    \caption{Supplementary qualitative results on the ScanNet200 and nuScenes datasets.}
    \label{fig:a}
\end{figure}

\end{document}